\documentclass[10pt, a4paper]{article}

\usepackage[]{lrec-coling2024} 

\usepackage{subcaption}
\usepackage{graphicx}

\usepackage{amsfonts}
\usepackage{amsmath}

\title{Vygotsky Distance: Measure for Benchmark Task Similarity}

\name{Maxim K. Surkov, Ivan P. Yamshchikov} 

\address{LEYA Laboratory, Higher School of Economics \& ITMO University, St. Petersburg\\ CAIRO, Technical University of Applied Sciences Würzburg-Schweinfurt \\
ivan.yamshchikov@thws.de}

\abstract{
Evaluation plays a significant role in modern natural language processing. Most modern NLP benchmarks consist of arbitrary sets of tasks that neither guarantee any generalization potential for the model once applied outside the test set nor try to minimize the resource consumption needed for model evaluation. This paper presents a theoretical instrument and a practical algorithm to calculate similarity between benchmark tasks, we call this similarity measure "Vygotsky distance". The core idea of this similarity measure is that it is based on relative performance of the "students" on a given task, rather that on the properties of the task itself. If two tasks are close to each other in terms of Vygotsky distance the models tend to have similar relative performance on them. Thus knowing Vygotsky distance between tasks one can significantly reduce the number of evaluation tasks while maintaining a high validation quality. Experiments on various benchmarks, including GLUE, SuperGLUE, CLUE, and  RussianSuperGLUE, demonstrate that a vast majority of NLP benchmarks could be at least $40\%$ smaller in terms of the tasks included. Most importantly, Vygotsky distance  could also be used for the validation of new tasks thus increasing the generalization potential of the future NLP models.
 \\ \newline  }

\begin{document}

\maketitleabstract

\section{Introduction}

Increasingly large language models such as \cite{anil2023palm,xu2023wizardlm,taori2023alpaca, vicuna2023, chowdhery2022palm, bajaj2022metro, zoph2022designing, raffel2020exploring, brown2020language,devlin-etal-2019-bert} manifest a clear trend to develop large, foundational models and then fine-tune in on a variety of NLP tasks depending on the use case. To prove that the new proposed model is beating the state-of-the-art solution, one typically evaluates the model on a series of tasks on which the suggested method is to be better than the existing ones. The community is standardly using conventional benchmarks to see if a given model is superior to the previous ones. When one talks about industrial applications of natural language processing systems, benchmarking becomes even more critical since, in practice, one has to balance various aspects of the systems, such as speed, accuracy, interpretability, etc.

Moreover, both in industry and academia, one is often interested in assessing the model's generalizing potential. The current approach is somewhat extensive, namely, 'the more tasks we use to evaluate, the better'. However, this is impractical and can be computationally costly. More importantly, such an approach diverts the attention of the NLP community from research into new qualitative and quantitative methods that could rigorously measure generalization potential. This paper suggests an ad-hoc approach to benchmark task similarity evaluation that could add rigour to NLP evaluation. We hope that such an assessment method will stimulate the search for broader sets of tasks that could endow the models with higher generalization capacity.


We regard {\it a benchmark} as a set of tasks, metrics, and evaluation methodology (usually, this is some form of aggregation over the included tasks). Each task is typically a dataset consisting of several samples of input texts and target output results. To properly evaluate the model, one typically fits the model on the training subset and then calculates metrics values on the evaluation part. Finally, one uses some aggregation method to get the model's final score that reflects the evaluated approach's quality on the given benchmark. For example, one of the most popular benchmarks in the field of natural language processing (NLP) is the General Language Understanding benchmark (GLUE) \cite{wang2018glue}, which was further extended in SuperGLUE \cite{wang2019superglue}. This benchmark has analogs in other languages: CLUE \cite{xu2020clue} for Chinese, or RussianSuperGLUE \cite{shavrina2020russiansuperglue} for Russian. All of them consist of about $10$ tasks. To get the final score of the model, one typically takes the average accuracy among all the problems provided in a given benchmark.

In this paper, we introduce the notion of Vygotsky distance — a measure of task similarity evaluated with respect to the relative ranking of the models on the given task. We demonstrate that most benchmarks contain up to fifty percent of the tasks that could be regarded as redundant. We show that removing these tasks from the benchmark has virtually no effect on the resulting assessment of the models' generalization capabilities. We run the experiments using evaluation results of all the NLP models on all available benchmarks provided by Papers With Code\footnote{\url{https://paperswithcode.com/}}. We mainly focus on GLUE, SuperGLUE, CLUE, and RussianSuperGLUE. The most valuable contribution of the paper is that the proposed benchmark compression method could be used to evaluate new benchmarks. The further a new benchmark is from the available tasks in terms of the Vygotsky distance the more value it carries. 

Section \ref{sec:benchmarks_graph_representation} is devoted to the graph representation technique that we developed to represent existing benchmarks. We consider each separate task in the benchmark as the vertex. We build an edge between each pair of tasks and assign a weight to it. The weight depends on the difference in the models' performance on the two tasks. This graph allows us to analyze the entire benchmark and retrieve the most meaningful tasks. Using the formalism we have proposed, we can determine which tasks in the benchmark are almost identical in terms of model evaluation and which are not. Further theoretical development of this framework is beyond the scope of this article, but this practical mechanism of benchmark compression is described in Section~\ref{sec:benchmarks_compression}. Suppose one splits the entire benchmark into public and private non-empty disjoint subsets. In that case, they could train classifiers to predict how models compare on the private leaderboard using the results obtained on the public leaderboard. Moreover, one can also predict the exact models' scores. This way, one could implicitly estimate the amount of information about the private subset of tasks available in the public part of the benchmark. For private leaderboard predictions, we use Support Vector Machine \cite{cortes1995support}, Gaussian Process \cite{williams2006gaussian}, and Multilayer Perceptron \cite{bishop1995neural}. Equipped with the collected data, we show how one can select a small subset of tasks within every benchmark so that it would suffice to predict the information of the models on the rest of the tasks.

The main contributions of this paper are as follows:
\begin{itemize}
    \item we propose a novel way to analyze benchmarks and evaluation systems representing them as weighted undirected graphs, which allows us to retrieve non-trivial structural properties of benchmarks and assess the similarity between the tasks in a given benchmark;
    \item we develop a framework that allows us to evaluate and compare NLP models on a small subset of benchmark tasks with the aforementioned Ad-Hoc benchmarks compression mechanism. This method significantly reduces the resources required for the qualitative evaluation of large NLP models;
    \item we conduct extensive analysis of the GLUE \cite{wang2018glue}, SuperGLUE \cite{wang2019superglue}, CLUE \cite{xu2020clue}, RussianSuperGLUE \cite{shavrina2020russiansuperglue} using our approach.
\item we suggest that further tasks that extend NLP benchmarks are to be analyzed in terms of Vygotsky distance to ensure that they significantly differ from the existing NLP tasks.
\end{itemize}   

\section{Benchmarks Graph Representation}
\label{sec:benchmarks_graph_representation}

\begin{figure*}[t]
    \centering
	\includegraphics[scale=0.35]{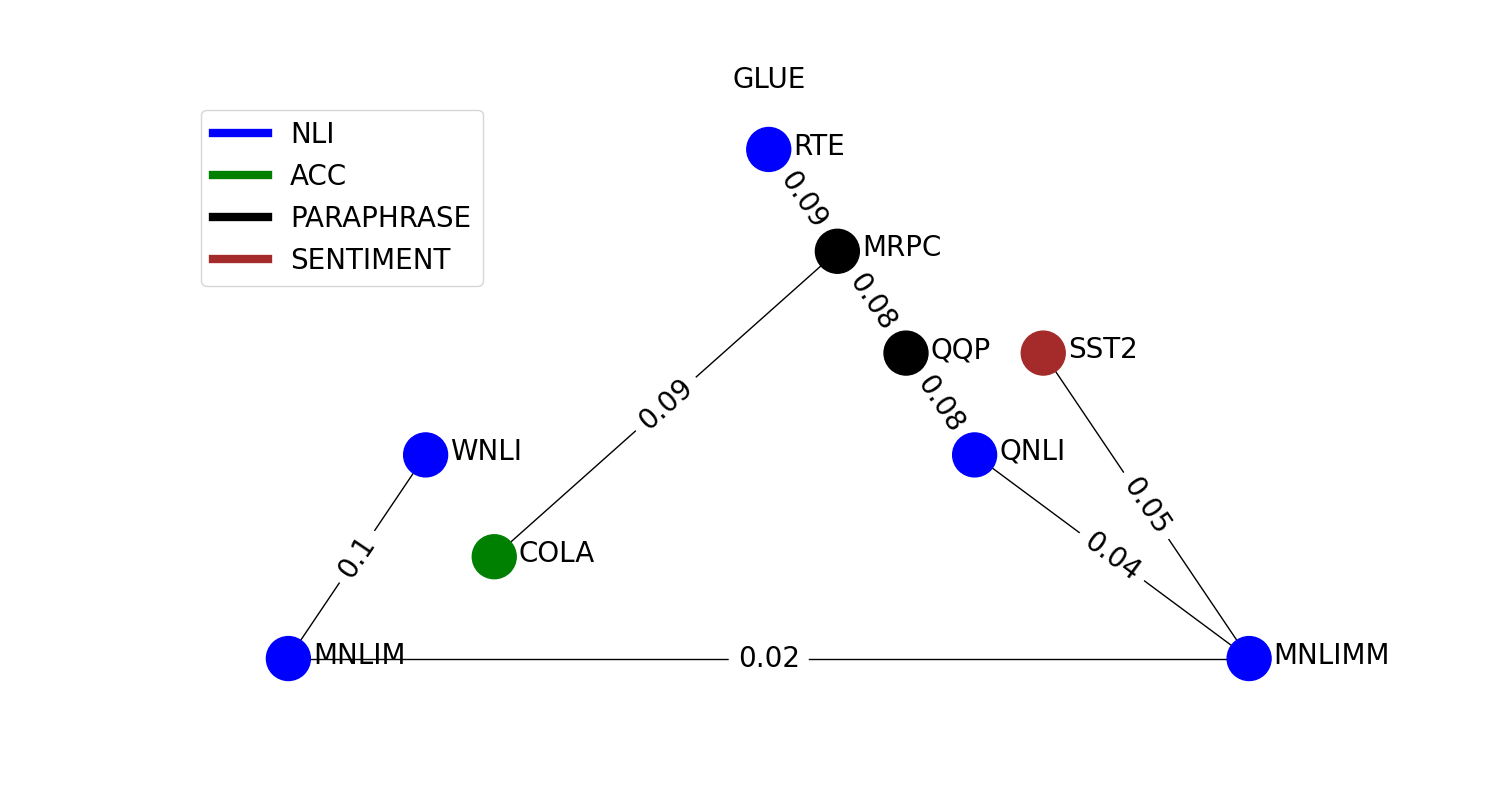}
	\caption{Minimum Weight Spanning Tree of the GLUE benchmark.}
	\label{fig:mst_glue}
\end{figure*}

\begin{figure*}[t]
    \centering
	\includegraphics[scale=0.35]{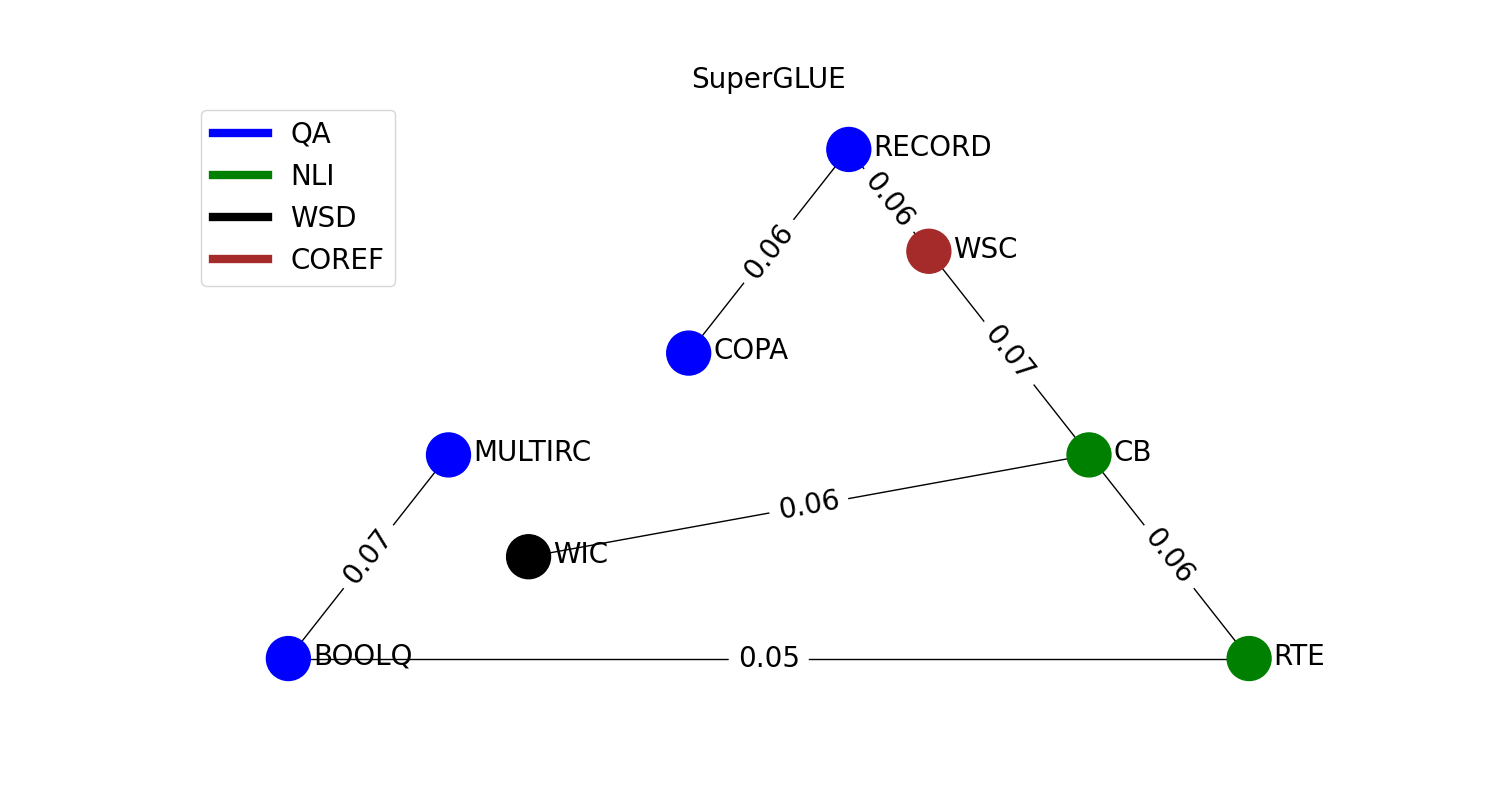}
	\caption{Minimum Weight Spanning Tree of the SuperGLUE benchmark.}
	\label{fig:mst_superglue}
\end{figure*}

Each task has a sorted list of models according to their score on a considered dataset. Thus we can think about each task as a permutation of evaluated models. As a result, we have a set of permutations to work with. Let us define a natural weight function that evaluates the measure of similarity of two permutations:
\begin{equation} \label{eq:V}
    w(\pi,\sigma) = inv(\pi\circ \sigma^{-1})
\end{equation}

where $\pi$ and $\sigma$ are two permutations of model rankings, and $inv$ is the function that calculates the number of inversions in the permutation:
$$inv(\pi) = |\{(i,j)\colon i < j, \pi_i > \pi_j\}|$$
In short, $w(\pi,\sigma)$ is the number of pairs of two models such that the first model is better than the second one on the task corresponding to permutation $\pi$, but the second model is better than the first one on the task corresponding to permutation $\sigma$. Then we can consider each task as the vertex of the complete graph where for each pair of vertices $(u, v)$ there is an edge of weight $w(u, v)$ scaled to $[0, 1]$. Thus we have an undirected weighted graph with non-negative weights that comply with the triangle inequality. The proof that $w$ satisfies the conditions of a metric space is in the Appendix~\ref{a:metric_space_proof}. We suggest to call distance defined in Equation \ref{eq:V} \emph{Vygotsky distance}.

Typically in NLP we use benchmark tasks to evaluate models, but how could one evaluate the similarity of tasks themselves? It is intuitively clear that formal comparison of datasets is futile. After all, the similarity of datasets only matters in the context of the current state of the art in the field at large. Here we suggest a task similarity measure that is based on the performance of the models instead. This "learner-first" approach inspired us to call the proposed similarity metric after a psychologist Lev Vygotsky who introduced the notion of the "zone of proximal development" \cite{semyonovich1978mind}. It represents "the distance between the actual development level as determined by independent problem solving and the level of potential development as determined through problem solving under adult guidance or in 
 collaboration with more capable peer" \cite{semyonovich1978mind}. We suggest to adopt this idea of a distance between tasks defined in context of the learners relative performance as a measure of task similarity and suggest a formal definition for such distance.

Relying on the metric space property of the Vygotsky distance, one can further apply a broad range of techniques for graph analysis (e.g., spectral clustering or topological data analysis). These research directions are out of the scope of this article. Instead, we focus on the practical aspects of the obtained graph representations for NLP benchmarks. 

Let us consider the minimum weight spanning tree of the benchmark graph. This entity (MST) has a series of useful properties that help us to understand the structure of the benchmarks more clearly:

\begin{itemize}
    \item MST is a planar graph, and we can plot it on the chart without edge intersections leading to clear graphic representation;
    \item for each pair of vertices $(u, v)$ there is a lower bound of $w_l(u, v)$ which is equal to the maximal weight of all edges on the path between $u$ and $v$ in the tree (otherwise, we can replace edge with the maximal weight on the path with an edge $(u, v)$ and get the tree with strictly less total weight);
    \item for each pair of vertices $(u, v)$ there is an upper bound of $w_u(u, v)$ which is equal to the sum of weights of all edges on the path between $u$ and $v$ in the tree (according to the metric space condition).
\end{itemize}

For better understanding, we address the reader to Figures~\ref{fig:mst_glue} -- \ref{fig:mst_superglue} (more MST plots can be found in Appendix~\ref{a:mst_of_nlp_benchmarks}). Let us consider an example of the edge between COLA and MNLIMM tasks of the GLUE benchmark. This edge weights $0.11$, which is bigger than the maximum weight among the edges on the path in the tree and smaller than the sum of all the weights on the path equal to $0.09 + 0.08 + 0.08 + 0.04 = 0.29$. As a result, MST gives us a visual representation of the whole benchmark, and we can easily estimate how far two benchmarks are in terms of the models' ranking on them. Moreover, we can notice some meaningful geometric properties, one of which is the so-called convexity. Notice that we can select groups of tasks in the benchmark which are close to each other (e.g., WNLI, MNLIM, MNLIMM, QNLI) and belong to the same problem type (e.g., Natural Language Inference (NLI)). One sees convexity in many benchmarks (see Appendix~\ref{a:mst_of_nlp_benchmarks}). However, there could be 'exceptional' tasks (e.g., Recognizing Textual Entailment (RTE)) too far from their groups. Reasonable research questions naturally follow from this fact:
\begin{itemize}
    \item \textit{How do we determine the type of tasks?}
    Currently, the typology of NLP tasks is based on the 'gut feeling' of the authors that develop a dataset. However, our method allows for a more rigorous answer to this question. \textit{One can either find the nearest group of already existing tasks and assume that the new one is of the same type, or if there is no such group, one might, indeed, claim that the task is of a novel type}. 
\item \textit{Why are there tasks that are far from their groups, and what distinguishes them from the rest of the group?}
One can immediately see some tasks that are classified as similar by the researchers that proposed them yet turn far apart in terms of the Vygotsky distance. For example, Figure~\ref{fig:mst_glue}  shows that some tasks that humans consider as NLI tasks could be further apart from each other than from sentiment. This means that some internal properties of the tasks make them 'look' different from the models' perspective. However, these differences are not evident to humans. Yet finding these differences might be insightful for further NLP research.
\end{itemize}

Another critical aspect of the proposed graph representation of a benchmark is that further contributions naturally follow from it. Using this formalism, we can easily describe which tasks are different and which are not. Moreover, the above approach gives rise to the idea that there are a lot of almost identical tasks. Thus, one can select a smaller subset of problems to be used in the qualitative models' evaluation benchmark. One could also quantify the potential for the generalization abilities of the model.

\section{Benchmarks Compression}
\label{sec:benchmarks_compression}

This section shows how the graph representations of benchmarks described above open up new possibilities for a benchmark compression algorithm.

\begin{figure}[t]
    \centering
	\includegraphics[scale=0.22]{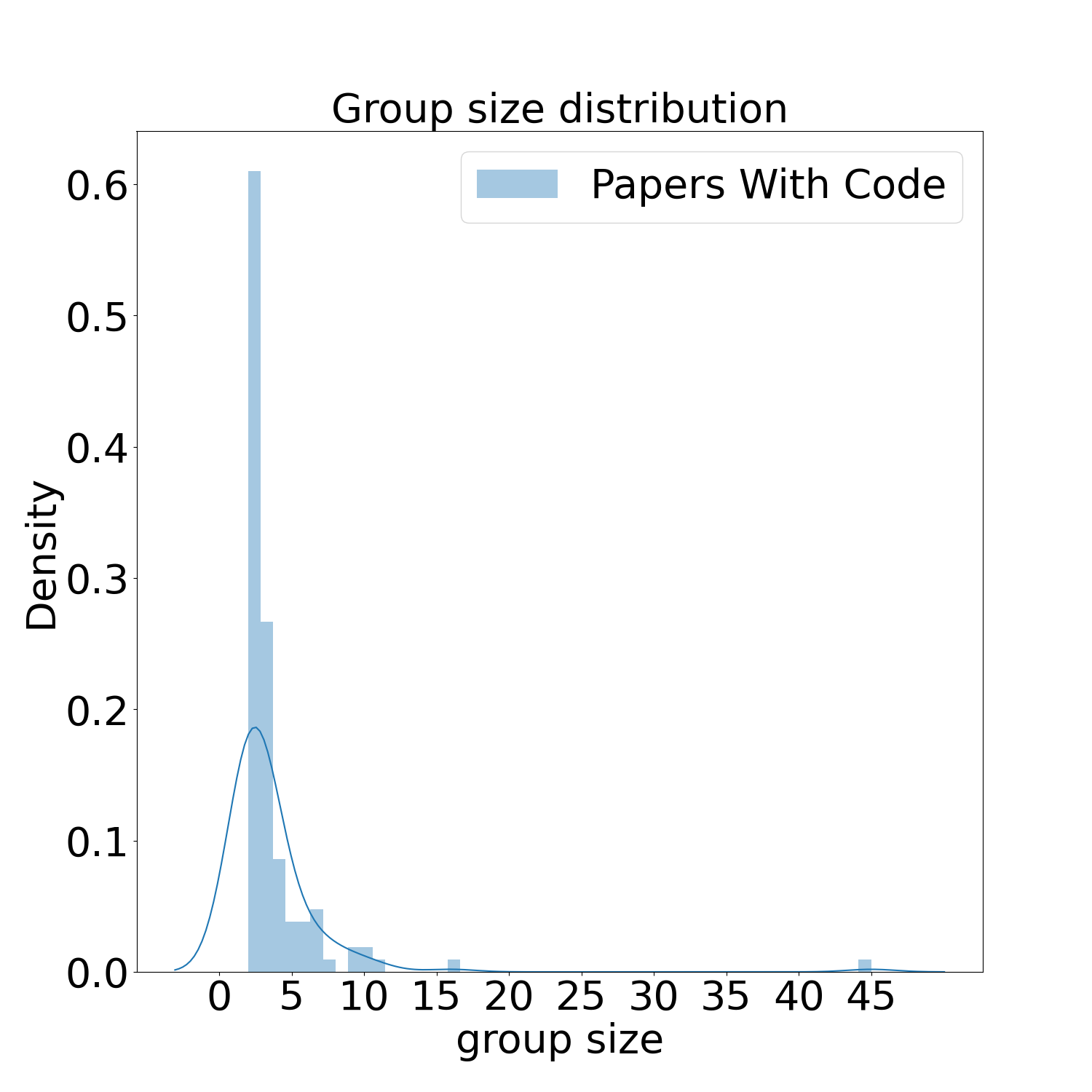}
	\caption{Distribution of benchmark sizes among the entire "Papers With Code" database.}
	\label{fig:all_sz_distr}
\end{figure}

\begin{figure*}[t]
    \centering
    \includegraphics[scale=0.27]{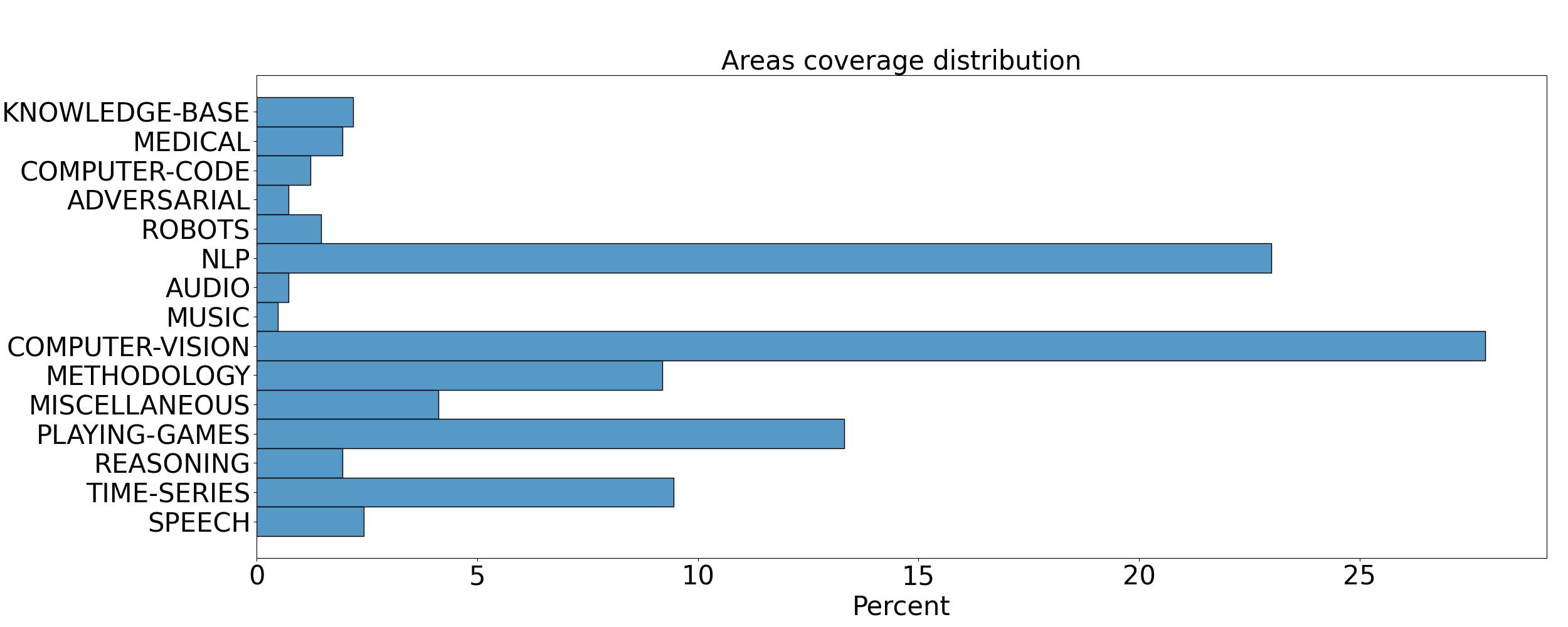}
	\caption{Machine learning areas coverage distribution where the contribution of each field is calculated as the total number of tasks.}
	\label{fig:areas_coverage_hist}
\end{figure*}

\begin{subsection}{Benchmarks structure and representations}

Before further details regarding the compression algorithm, we first formally describe the collected data. Each task has the list of models' evaluation results, namely the list of records that contains the name of the model (e.g., "BERT") and the set of metrics (e.g., "accuracy" is equal to $0.95$, "f1" is equal to $0.88$) computed on the evaluation part of the dataset related to the task. We select groups of tasks in such a way that the following conditions are satisfied:

\begin{itemize}
    \item There are at least $C$ models that have been tested on all the tasks in the group, where $C$ is the constant that depends on the benchmark (on average, $C\approx 10$).
    \item There is a real-valued metric that was used to evaluate the models.
\end{itemize}

We process the entire "Papers With Code," including all known areas in machine learning. As a result, we collect $120$ task groups of different sizes distributed as shown in Figure~\ref{fig:all_sz_distr}. It turned out that NLP is the broadest area in "Papers With Code" after Computer Vision among $16$ machine learning fields and covers $23\%$ of the database (Figure~\ref{fig:areas_coverage_hist}). 

Each group can be considered as the standalone benchmark and analyzed independently. Each benchmark is the table of real values which consists of the $n_{models}$ rows, each of which corresponds to the particular model, and $n_{tasks}$ columns, each of which corresponds to the particular task (dataset) in the benchmark. Let us call this table "the leaderboard" and denote as the $M \in \mathbb{R}^{n_{models}\times n_{tasks}}$, where $M_{i,j}$ is the score of the $i$-th model on the $j$-th task in the zero-based enumeration.

\end{subsection}

\begin{subsection}{Compression Algorithm}

There are $(2^{n_{tasks}} - 2)$ ways to split the entire set of tasks into two non-empty disjoint subsets: public and private leaderboards. We are interested in estimating the amount of information about the private part contained in the public part. The explicit calculation seems almost impossible, but we can evaluate it implicitly. To do that, we formulate two types of prediction problems:

\begin{itemize}
    \item Models comparison. We have scores of two particular models on the tasks which belong to the public part. According to the average model score, the problem is to predict which of the two given models will be better on the private leaderboard;
    \item Model's score estimation. The problem is to predict the exact model average score on the private leaderboard, given information on the public leaderboard.
\end{itemize}

\begin{figure}[t]
    \centering
    \includegraphics[scale=0.21]{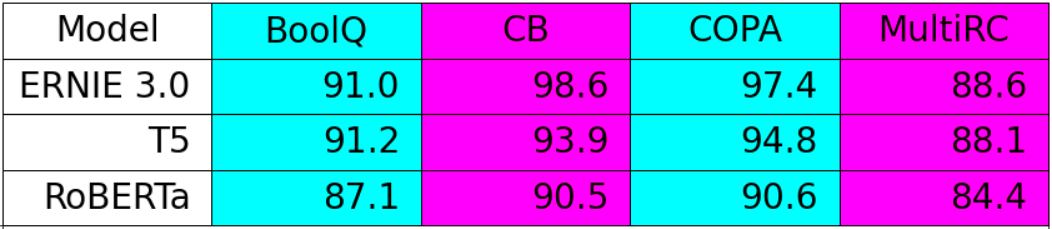}
	\caption{Example of splitting SuperGLUE into public and private leaderboards. Cyan-colored columns (BoolQ, COPA) belong to the public subset, magenta-colored columns (CB, MultiRC) belong to the private subset.}
	\label{fig:super_glue_leaderboard_example}
\end{figure}

One can solve both problems using existing classification and regression models. Fixing the number of tasks we are ready to use in a public part, one can then find the best splitting with respect to the prediction model accuracy. Alternatively, one can fix the accuracy threshold. Then one can find a public leaderboard with a minimal number of tasks allowing one to predict information about the private leaderboard with the desired accuracy. 

\textbf{Models Comparison}\\
First of all, we have to prepare the data for classification. Suppose that we fix the public $M^{pub}~\in~\mathbb{R}^{n_{models}\times n_{tasks}^{public}}$ and the private $M^{pr}~\in~\mathbb{R}^{n_{models}\times n_{tasks}^{private}}$ leaderboards (Figure~\ref{fig:super_glue_leaderboard_example}). We can define the \textit{compression} as the $n_{tasks}^{public}\big/n_{tasks}$. Let us denote the function:
$$pairs(A) = \{(i, j)|i,j \in A \texttt{ and } i < j\}$$
Note that $|pairs(A)| = \Theta(|A|^2)$.
To avoid data leakage during evaluation, we split the entire set of models (rows of the leaderboards) into two disjoint subsets 
$$R_{train} \sqcup R_{val} = \{0, 1, \ldots, n_{models} - 1\}$$
Then we consider sets of pairs 
$$P_{tr} = pairs(R_{train}), P_{val} = pairs(R_{val})$$

Finally, we can construct the dataset for the classification problem in the following way:
$$X_{tr} = \{\texttt{concat}(M^{pub}_i, M^{pub}_j)|(i,j)\in P_{tr}\}$$
$$y_{tr} = \left\{\mathbb{1}\left[{\texttt{avg}(M_i^{pr})<\texttt{avg}(M_j^{pr})}\right]|(i,j)\in P_{tr}\right\}$$
where \texttt{concat} is the concatenation of two vectors, \texttt{avg} is the mean value of the vector. Similarly, $X_{val},y_{val}$ are built. In this way, input data is the set of vector concatenations among all possible pairs of rows in the public leaderboard. Using this information, the prediction model determines whether the first model in the pair has a lower average score than the second one on the private leaderboard, which is a standard binary classification problem.

\textbf{Model's Score Estimation}\\
Let us define the following sets:
$$X_{tr}=\{M_i^{pub}|i \in R_{tr}\}$$
$$y_{tr}=\{\texttt{avg}(M_i^{pr})|i\in R_{tr}\}$$
$$X_{val}=\{M_i^{pub}|i \in R_{val}\}$$
$$y_{val}=\{\texttt{avg}(M_i^{pr})|i\in R_{val}\}$$

In short, the input data is the set of public score vectors, and the target data is the set of average private scores.
Notice that the datasets constructed in this way are suitable for the standard regression problem.

\end{subsection}

\section{Experiments}

\begin{figure}
    \centering
	\includegraphics[scale=0.18]{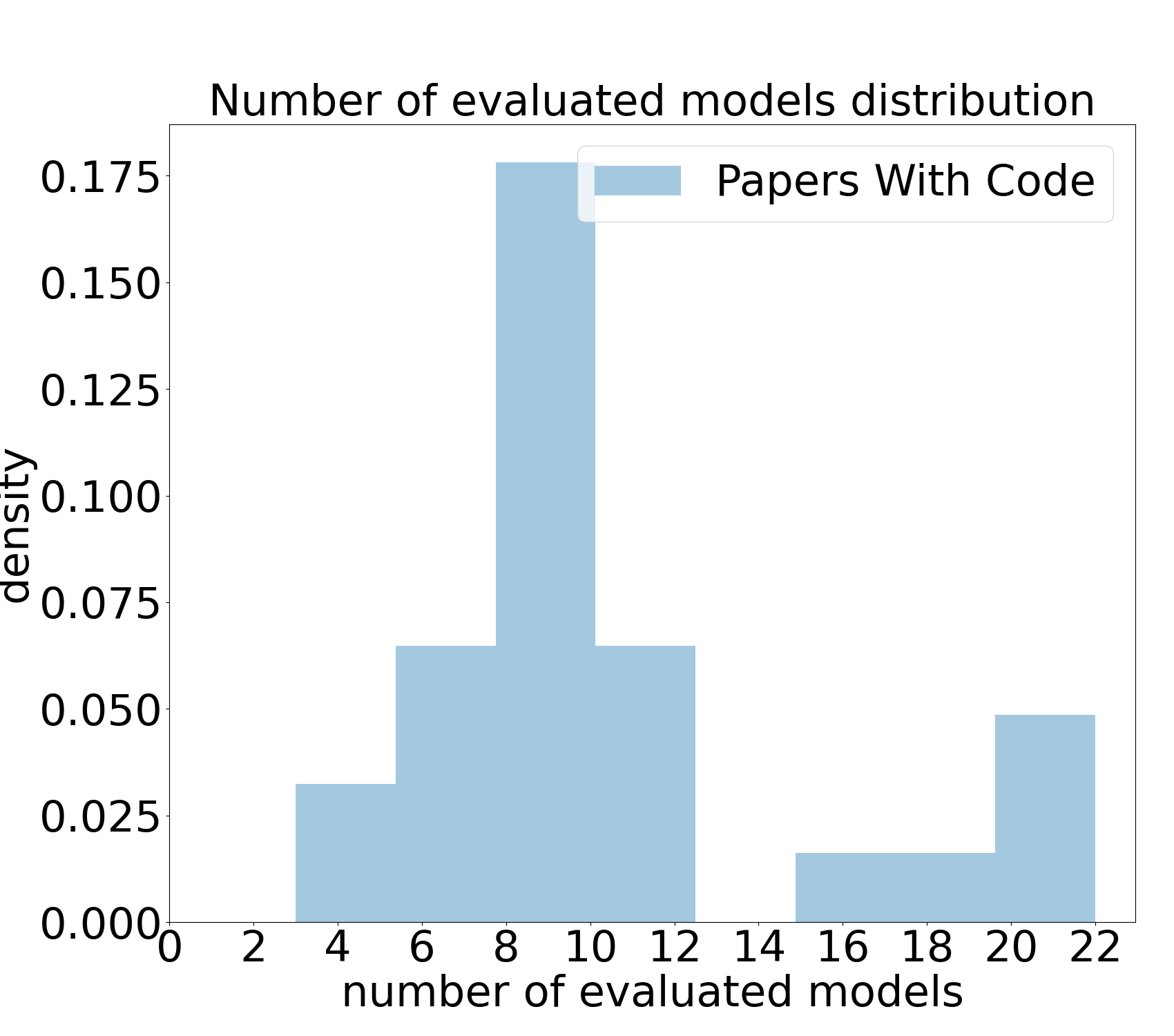}
	\caption{Distribution of a number of evaluated models among all the NLP benchmarks in the entire Papers With Code.}
	\label{fig:nlp_n_models_distr}
\end{figure}

\begin{table}[t]
	\centering
	\begin{tabular}{l|c}
	    \hline
	    Benchmark & $n_{models}$ \\
	    \hline
	    Papers With Code & $3-22$ \\
	    GLUE & $193$ \\
	    SuperGLUE & $34$ \\
	    RussianSuperGLUE & $23$ \\
        CLUE & $12$ \\
        \hline
	\end{tabular}
     \caption{The number of models evaluated on the benchmark.}
	\label{table:num_of_models_in_benchmark}
\end{table}


\begin{subsection}{Data}
In our experiments, we use the data of GLUE, SuperGLUE, CLUE, RussianSuperGLUE, and all the benchmarks in Papers With Code which belong to the NLP field. The number of models which was evaluated depends on the benchmark and is distributed as shown in Figure~\ref{fig:nlp_n_models_distr} and Table~\ref{table:num_of_models_in_benchmark}. For the model's score estimation, data were normalized so that all target scores are in $[0, 1]$. We have $30$ benchmarks and about $2600$ divisions on public and private leaderboards for analysis.
\end{subsection}

\begin{subsection}{Prediction Models}
We use Support Vector Machine (SVM), Gaussian Process Model (GP), and Multilayer Perceptron (MLP) for both the classification and regressions problems with default hyperparameters:
\begin{itemize}
    \item SVM: standard support vector machine with the radial basis function kernel and $l_2$ regularization with $C=1$.
    \item GP: Gaussian process with the Laplace approximation of non-gaussian posterior with the radial basis function kernel.
    \item MLP: neural network of depth $4$ layers with hidden size of $16$ and ReLU activation function. The model is trained using Adam optimizer with a learning rate of $10^{-3}$ and batch size of $32$ for $10$ epochs.
\end{itemize}
\end{subsection}

\begin{subsection}{Evaluation Metrics}
We use accuracy, F1-score, precision, recall, and ROC-AUC metrics for a classification problem. We use MSE, RMSE, MAE, MAX-ERROR, and R2-score metrics for a regression problem.
\end{subsection}

\begin{figure*}[t]
    \centering
    \begin{subfigure}{\textwidth}
        \centering
    	\includegraphics[scale=0.25]{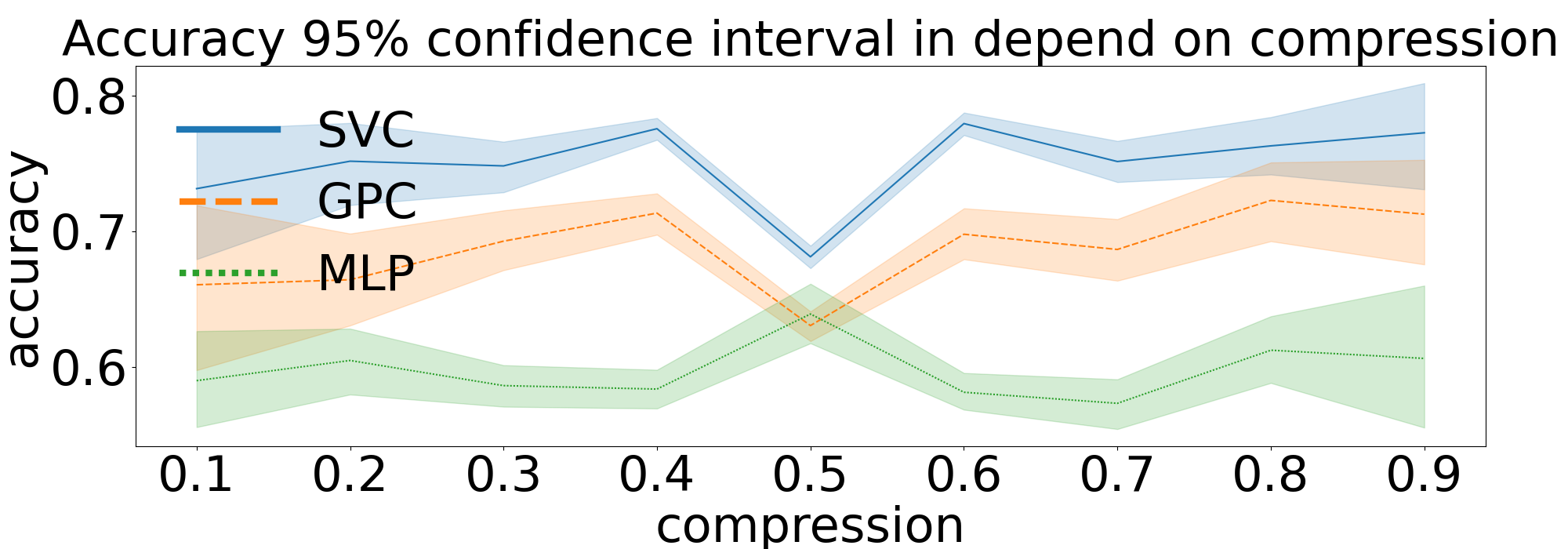}
    	\caption{Accuracy dependency}
    	\label{fig:all_acc_95_conf}
    \end{subfigure}
    \begin{subfigure}{\textwidth}
        \centering
    	\includegraphics[scale=0.25]{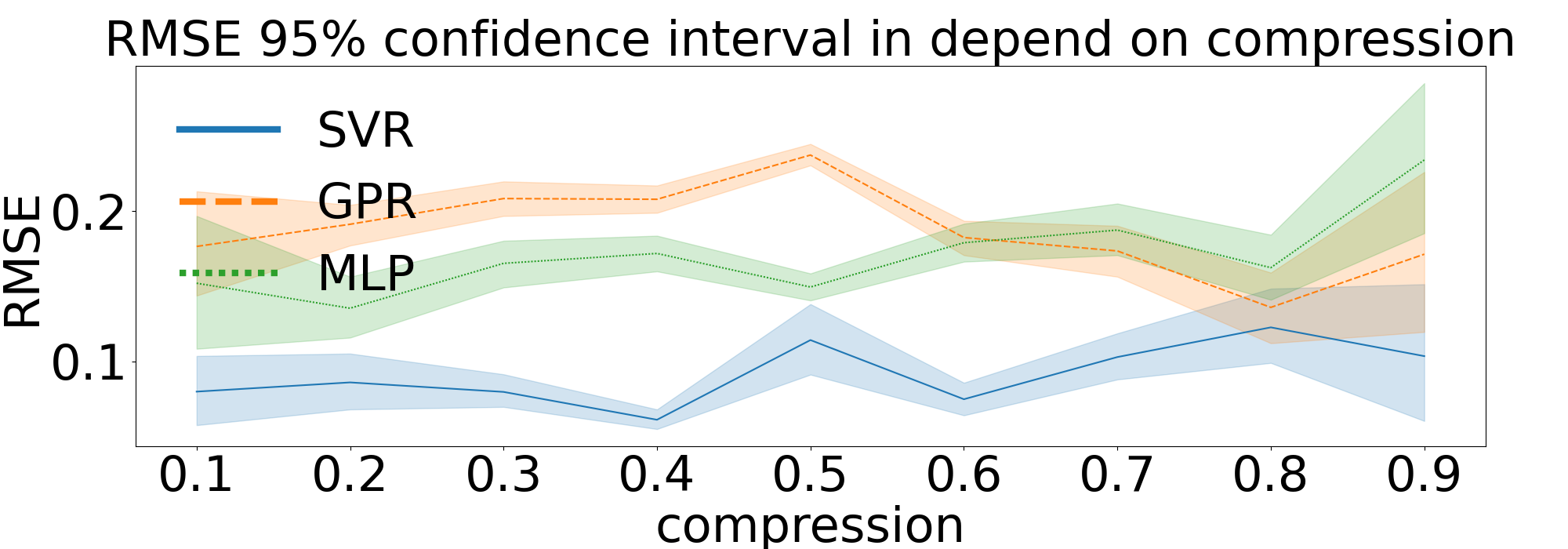}
    	\caption{RMSE dependency}
    	\label{fig:all_rmse_95_conf}
    \end{subfigure}
    \caption{Metric value $95\%$ confidence interval in depend on the compression rate among all the public subsets of all the benchmarks.}
\end{figure*}

\begin{subsection}{Results}

\begin{table*}[t]
	\centering
	\begin{tabular}{l|ccccc||ccccc}
	    \hline
	    Model & & \multicolumn{3}{c}{Classification} & & & \multicolumn{3}{c}{Regression}  & \\
	    \hline
	    \hline
	    & acc & F1 & prec & recall & ROC-AUC & MSE & RMSE & MAE & MAX-ERR & R2 \\
	    \hline
	    SVM & 0.74 & 0.70 & 0.64 & 0.83 & 0.71 & \textbf{0.33} & \textbf{0.20} & \textbf{0.26} & \textbf{0.61} & \textbf{0.41} \\
	    \hline
	    GP  & \textbf{0.79} & 0.67 & 0.68 & 0.72 & \textbf{0.78} & 0.40 & 0.35 & 0.29 & 0.72 & 0.0 \\
	    \hline
	    MLP & 0.78 & \textbf{0.75} & \textbf{0.80} & \textbf{0.95} & 0.7 & 0.49 & 0.55 & 0.42 & 0.79 & 0.0 \\
	    \hline
	\end{tabular}
     \caption{Mean metric values of the prediction models for the best (according to the metric) benchmark splitting into public and private leaderboards with compression of at most $40\%$ among all the benchmarks.}
	\label{table:best_subsets}
\end{table*}

First of all, we need to understand the best quality of the prediction model we can get with the fixed compression rate. To do that, let us turn to Figures~\ref{fig:all_acc_95_conf}~and~\ref{fig:all_rmse_95_conf}, which show the desired information. We can see the $95\%$ confidence interval of the prediction models' quality metric taken among all the public leaderboards of a specific compression rate among all considered benchmarks for each compression rate. Notice that regardless of the compression (except $\textit{compression} = 0.5$), there is a predictor that can compare models on the private leaderboard according to the average score with an accuracy of at least~$80\%$. Similarly, we can see a regressor that can predict an exact model's score on the private leaderboard with RMSE of at most $0.05$. Also, we can notice interesting behavior of the metric value dependency on the compression rate. Graphs are almost symmetrical. In the case of small compression, we do not have enough features of the data samples, while in the case of high compression, the number of features is too large for a given size of training data. Both situations lead to the predictors being under fitted. Thus, we can declare that the optimal compression rate is somewhere around $40\%$. We can use this threshold to analyze the results from a different perspective. 

Consider each benchmark separately and find such a public leaderboard whose size is at most $40\%$ of the total number of tasks, while a prediction model (SVM, GP, or MLP) solves a classification or regression problem using this public leaderboard with the best quality. Collect the mean metric value among all considered benchmarks. The best classification model is the MLP which gives us a prediction accuracy of about $80\%$. On the other hand, the best regression model is the SVM which predicts the exact model's score on the private leaderboard with RMSE of $0.2$. The results of this experiment are shown in Table~\ref{table:best_subsets}.

Finally, Figure~\ref{fig:all_rmse_95_conf} shows a vast number of subsets on which SVR can estimate the exact score with an error of at most $5\%$. Thus, the regression problem is more difficult than expected; see Table~\ref{table:best_subsets}. At the same time, the models' comparison problem remains solvable in both approaches to compression.

\end{subsection}

\begin{figure*}[t]
    \centering
	\includegraphics[scale=0.35]{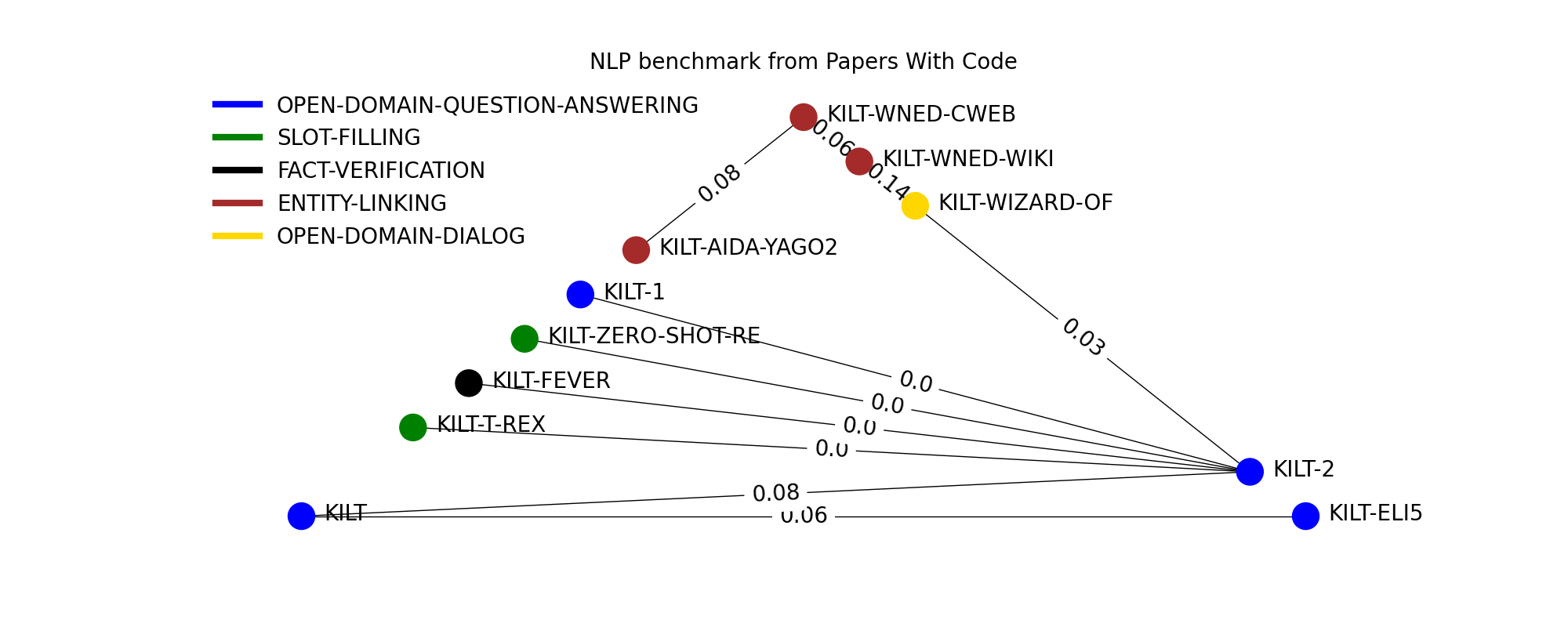}
	\caption{Minimum Weight Spanning Tree of the KILT benchmark.}
	\label{fig:mst_kilt}
\end{figure*}

\section{Discussion}
Our approach has several theoretical and practical implications. 

First, in addition to the questions and hypotheses set out in Sections~\ref{sec:benchmarks_graph_representation}~and~\ref{sec:benchmarks_compression}, there is another important observation for future research. The proposed graph representation for the evaluation benchmarks clearly emphasizes the areas that need more research community attention. See, for example, Figure~\ref{fig:mst_glue}. The "density" of the NLI group is quite small. This suggests that there is room for other NLI tasks and some other tasks that might be adjacent to NLI, say, some other areas of causal NLP. On the other hand, tasks that could be classified as forms of PARAPHRASE are located close to each other. The majority of paraphrase datasets that we have are far too similar in terms of Vygotsky distance. We either need some new paraphrase task that is radically different from the ones we have, or the whole problem of the paraphrase is well "covered" by the available benchmarks. Either way, the proposed graph representation is insightful for further research. 

Second, the proposed formalism could be used for prioritization of the tasks in terms of the generalization potential of the models. For instance, Figure~\ref{fig:mst_kilt} shows that SLOT-FILLING and FACT-VERIFICATION tasks are almost identical to OPEN-DOMAIN-QA. Indeed, if a model can answer the open questions well, then there should be no problems with slot filling or checking a fact for its correctness. However, inside the group of OPEN-DOMAIN-QA tasks, there are several datasets that differ significantly from each other. This concludes that question answering is, in some sense, a more complex family of tasks, the score on which is more important from the perspective of the generalization potential of a model.  

Finally, Vygotsky distance is useful for industrial applications. The best modern practices recommend using continuous testing systems, which means that after minimal changes in the project, one needs to run tests. Thus validation directly affects the product development time, which can be significantly reduced by applying the presented approach.

\section{Conclusion}
This paper presents a theoretical instrument for evaluation systems analysis representing benchmarks as undirected weighted graphs. We also introduce a metric for benchmark task similarity that we call Vygotsky distance. Using the proposed formalism, we demonstrate several meaningful properties of NLP benchmarks. 

In particular, we develop an algorithm for benchmark compression. Our approach allows us to analyze all the NLP benchmarks such as GLUE, SuperGLUE, CLUE, and RussianSuperGLUE and find out that we can keep only $40\%$ of the entire benchmark in such a way that there exists a prediction model which can estimate almost all remaining information with an accuracy of at least $80\%$ for Models Comparison, and with an error of $5$-$20\%$ for Exact Models' Score Estimating problem.

Finally, we discuss how the proposed metric of benchmark similarity can help to guide the development of new NLP benchmarks and improve generalization potential of the NLP models.

\section*{Limitations}

For this research, we used Papers with Code. We believe the results are reproducible with other sets of benchmarks, but this has not been proved within the paper. However we do not see any structural difficulties that would hinder the application of the proposed reasoning to different tasks.

\section*{Acknowledgements}
Mr. Surkov's work was supported by the grant for research centers in the field of AI provided by the Analytical Center for the Government of the Russian Federation (ACRF) in accordance with the agreement on the provision of subsidies (identifier of the agreement 000000D730321P5Q0002) and the agreement with HSE University No. 70-2021-00139

The authors are deeply thankful to Mr. Vladimir Smurygin and Prof. Dr. J{\"u}rgen Jost for deep insightful conversations and overall support of the project.


\nocite{*}
\section{Bibliographical References}\label{sec:reference}

\bibliographystyle{lrec-coling2024-natbib}
\bibliography{lrec-coling2024-example}


\appendix
\section{Metric Space Condition of Benchmark Graph Representation}
\label{a:metric_space_proof}

In Section~\ref{sec:benchmarks_graph_representation} we define the edge weight function:

$$w(\pi,\sigma) = inv(\pi\circ \sigma^{-1})$$

We have to prove that $w$ satisfies the conditions of a metric space.

The first condition is $w(u,v)\ge 0$ and $w(u,v)~=~0 \Leftrightarrow u = v$ (we assume that there are no absolutely identical tasks in terms of evaulated models permutation). The number of inversions is a non-negative value and 
$$w(u, u) = inv(u \circ u^{-1}) = inv(id) = 0$$

The second condition is a triangle inequality $w(x, y) \le w(x, z) + w(z, y)$. Notice that $w(x,y)$ is equal to the minimal number of swaps of two adjacent elements in a permutation needed to obtain permutation $y$ starting from the permutation $x$:
$$w(x,y) = n_{swaps}(x,y)$$

\begin{itemize}
    \item Notice that $w(x, y) = inv(x \circ y^{-1}) = |\{(i, j)\colon x_i < x_j, y_i > y_j\}|$.
    \item Let us prove that $w(x,y) \le n_{swaps}(x, y)$. Consider an optimal sequence of swapping two adjacent elements for transforming $x$ to $y$. Notice that each pair $(i, j)$ such that $x_i < x_j$ and $y_i > y_j$ should be fixed (both inequalities have to be of the same type). Therefore, there is a moment of time when this pair of elements will be on the adjacent positions, because after each operation each element's location changes by at most one, and at least one operation is required to fix the inversion.
    \item Let us prove that $w(x,y) \ge n_{swaps}(x, y)$. As long as $x \neq y$ there is $i$ such that $x_{i} < x_{i+1}$ and $y_{i} > y_{i+1}$. We can swap $x_{i}$ and $x_{i+1}$ decreasing $w(x, y)$ by one. In such a way we can obtain $x$ from $y$ by swapping two adjacent elements using exactly $w(x, y)$ operations which is an example, but probably not optimal. It gives us an upper bound for $n_{swaps}(x, y)$.
\end{itemize}

We can obtain $y$ from $x$ by transforming $x$ to $z$ and then $z$ to $y$. As a result, we present an example of swap sequence which allows us to obtain $y$ from $x$ using no more than $w(x, z) + w(z, y)$, but $w(x, y)$ is the minimal number of such swaps and we show that it is at most $w(x,z) + w(z,y)$.

\section{Minimum Weight Spanning Trees of NLP Benchmarks}
\label{a:mst_of_nlp_benchmarks}

\begin{figure}[h]
    \centering
	\includegraphics[scale=0.22]{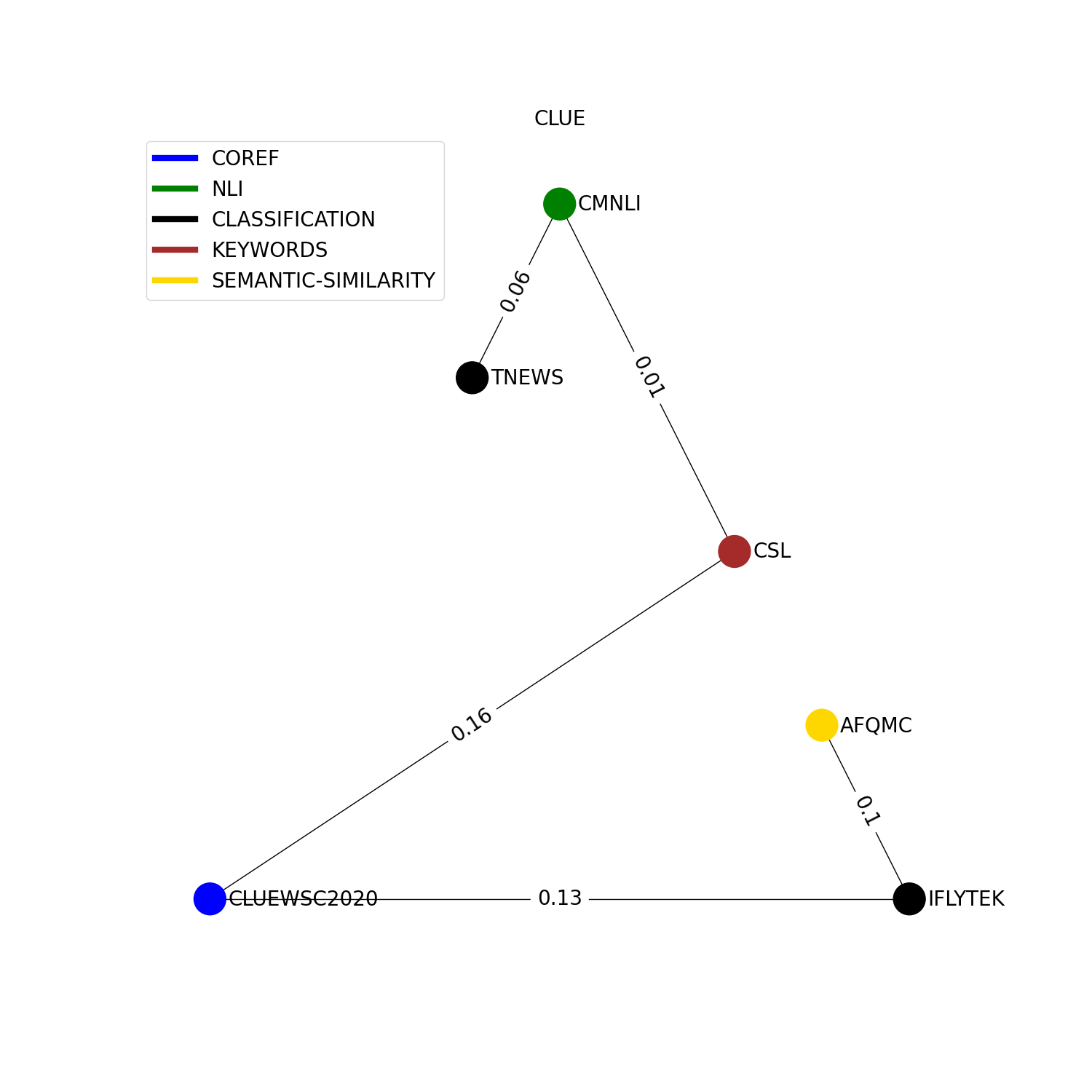}
\end{figure}


\begin{figure}[t]
    \centering
	\includegraphics[scale=0.22]{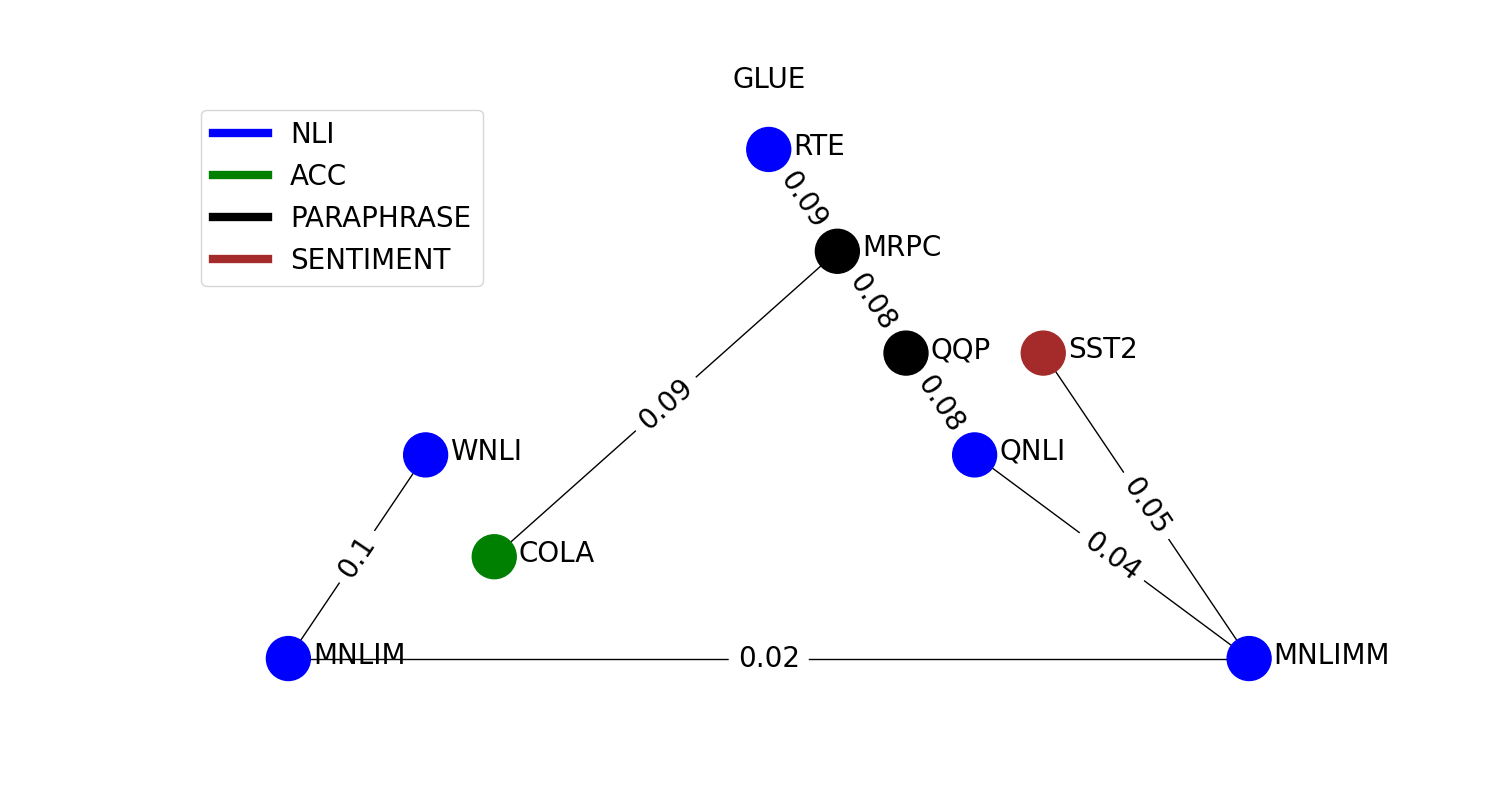}
\end{figure}

\begin{figure}[t]
    \centering
	\includegraphics[scale=0.25]{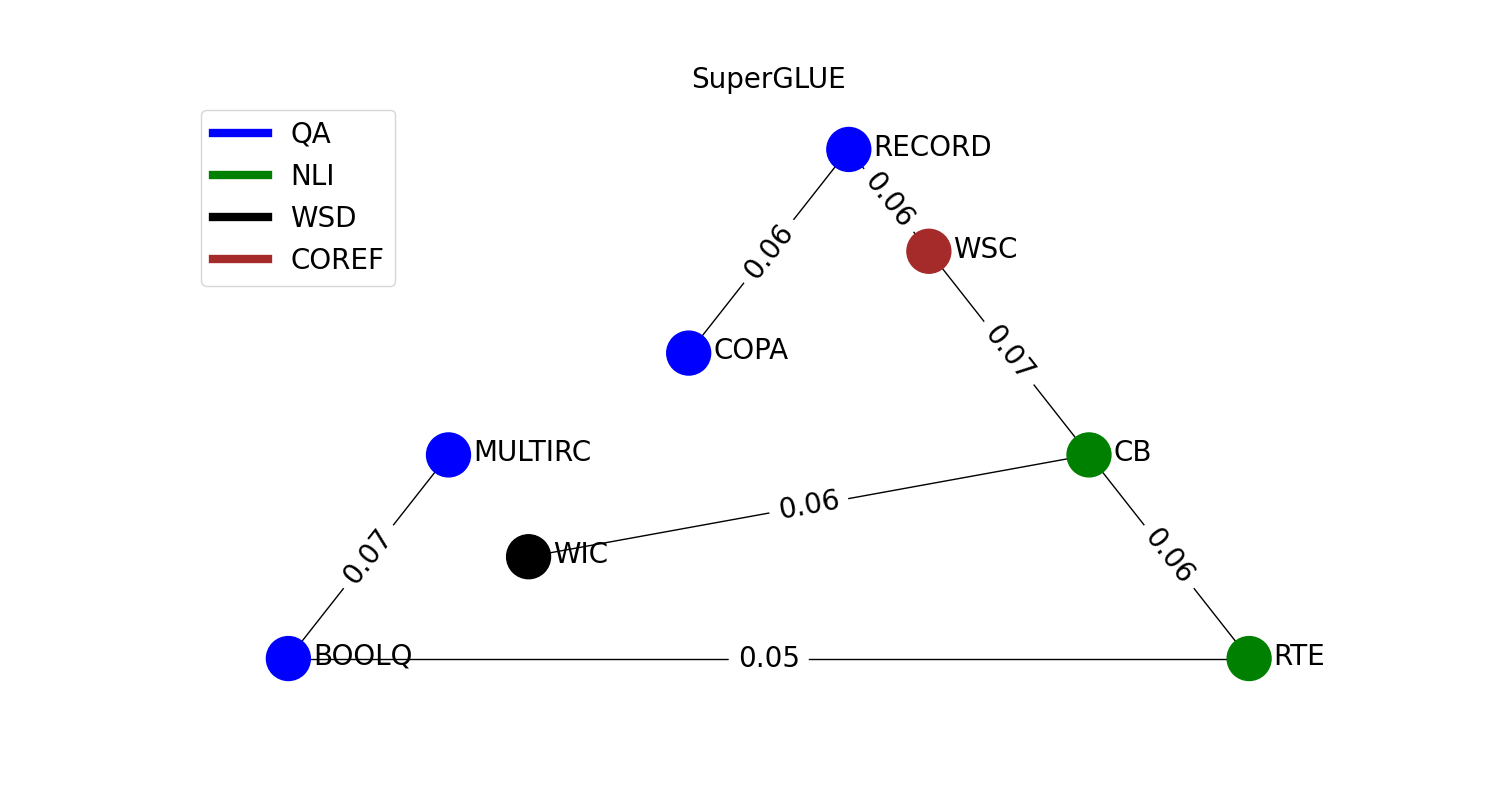}
\end{figure}

\end{document}